\documentclass[letterpaper, 10 pt, conference]{ieeeconf}  

\usepackage{graphicx,amsmath,amssymb,amsfonts,algorithm,algorithmic}
\usepackage{eqparbox,array,xcolor,cite}
\usepackage{adjustbox}

\IEEEoverridecommandlockouts                              
\title{\LARGE \bf
Towards High Efficient Long-horizon Planning with Expert-guided Motion-encoding Tree Search
}

\author{
Tong Zhou$^{1, \#}$,
Erli Lyu$^{2, \#}$, Jiaole Wang$^{3, *}$,
Guangdu Cen$^{3}$,  Ziqi Zha$^{3}$,\\
Senmao Qi$^{3}$, and Max Q.-H. Meng$^{4, *}$
\thanks{This work was supported in part by , and in part by .}
\thanks{$^{1}$ Tong Zhou is with Department of Electronic Engineering, The Chinese University of Hong Kong, Shatin, N.T., Hong Kong SAR, China.}
\thanks{$^{2}$ Erli Lyu is with Faculty of Applied Science, Macao Polytechnic University, Macao, China.}
\thanks{$^{3}$ Jiaole Wang, Guangdu Cen, Ziqi Zha, Senmao Qi are with School of Mechanical Engineering and Automation, Harbin Institute of Technology (Shenzhen), Shenzhen, China, 518055.}
\thanks{$^{4}$ Max Q.-H. Meng is with the Shenzhen Key Laboratory of Robotics Perception and Intelligence, and the Department of Electronic and Electrical Engineering, Southern University of Science and Technology, Shenzhen 518055, China, on leave from the Department of Electronic Engineering, The Chinese University of Hong Kong, Hong Kong, and also with the Shenzhen Research Institute of The Chinese University of Hong Kong, Shenzhen 518057, China.}
\thanks{$^*$ Corresponding authors: Jiaole Wang (e-mail: wangjiaole@hit.edu.cn),
Max Q.-H. Meng (e-mail: max.meng@ieee.org).}
\thanks{$^\#$ Tong Zhou and Erli Lyu contributed equally to this work.}
}

\begin{document}
\maketitle
\thispagestyle{empty}
\pagestyle{empty}

\begin{abstract}

Autonomous driving holds promise for increased safety, optimized traffic management, and a new level of convenience in transportation.
While model-based reinforcement learning approaches such as MuZero enables long-term planning, the exponentially increase of the number of search nodes as the tree goes deeper significantly effect the searching efficiency.
To deal with this problem, in this paper we proposed the expert-guided motion-encoding tree search (EMTS) algorithm.
EMTS extends the MuZero algorithm by representing possible motions with a comprehensive motion primitives latent space and incorporating expert policies to improve the searching efficiency.
The comprehensive motion primitives latent space enables EMTS to sample arbitrary trajectories instead of raw action to reduce the depth of the search tree.
And the incorporation of expert policies guided the search and training phases the EMTS algorithm to enable early convergence.

In the experiment section, the EMTS algorithm is compared with other four algorithms in three challenging scenarios.
The experiment result verifies the effectiveness and the searching efficiency of the proposed EMTS algorithm.

\end{abstract}

\section{INTRODUCTION}

Autonomous driving brings considerable potential to enhance safety, boost efficiency, and provide unprecedented convenience in transportation systems. However, it faces significant challenges such as complex decision-making processes, unpredictable real-world environments, interpretability issues.
Reinforcement Learning \cite{silver2016mastering,vinyals2019grandmaster,liu2021motor} (RL), with its capacity for adaptive learning and policy optimization, presents a robust approach to handle the complexity and uncertainty inherent in autonomous driving. 
Unlike rule-based methods, which often struggle to encompass the vast range of possible scenarios, RL learns interactively from the environment, adapting to various uncertainties and complexities. 


Model-based RL approaches like MuZero \cite{Schrittwieser2019MasteringAG}, which integrate MCTS, provide advantages for autonomous driving such as enabling long-term planning\cite{ozair2021vector}, enhancing robustness by modeling uncertainties\cite{antonoglou2021planning}, and increasing interpretability through traceable decision paths. 
However, the application of MCTS in autonomous driving poses two major challenges: computational complexity \cite{danihelka2021policy} due to the long-term search horizon, and sample efficiency problem\cite{ye2021mastering} for complex driving scenarios.

\begin{figure}[t]
    \centering
    \includegraphics[width=\columnwidth]{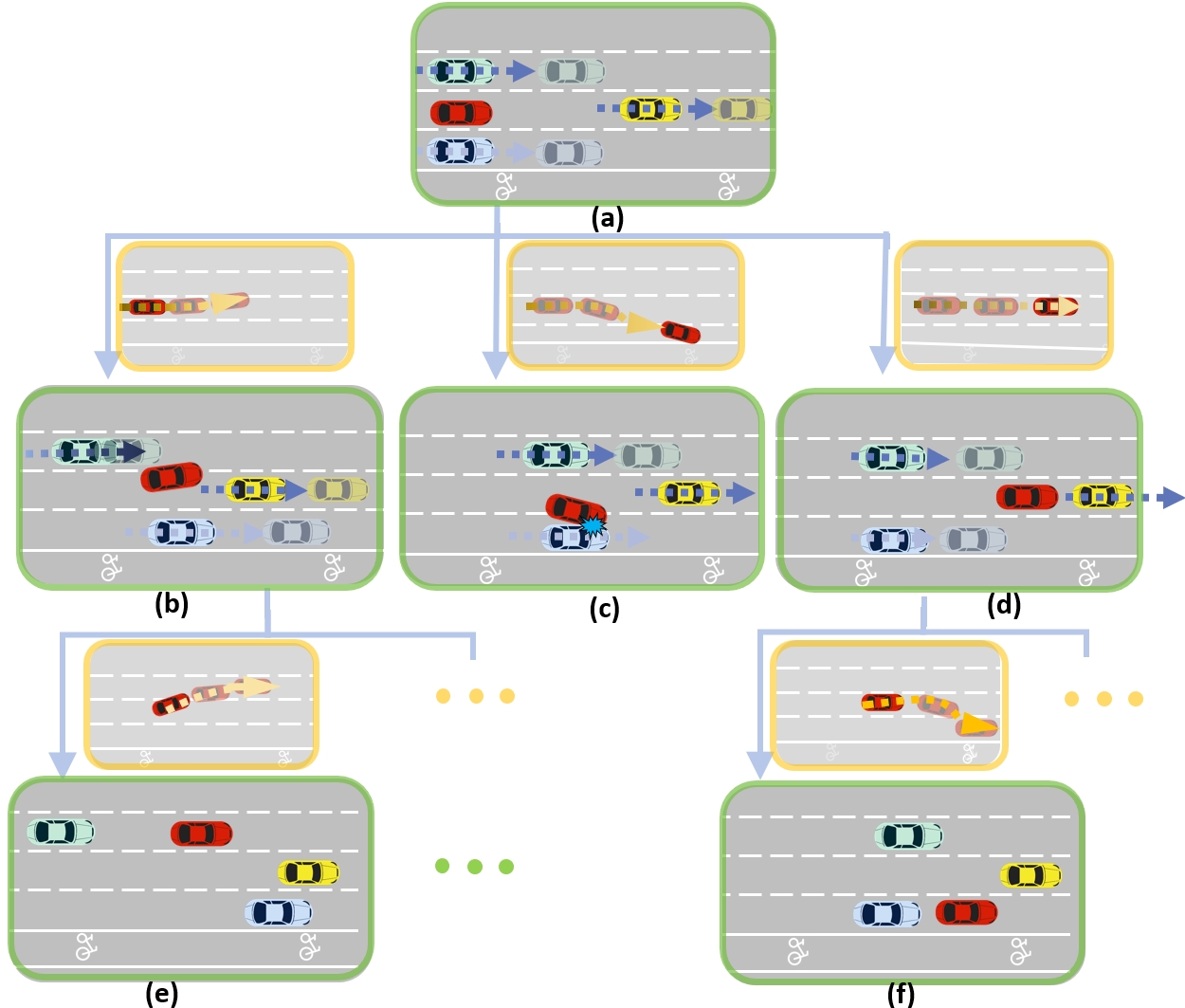}
    \caption{Illustration of Expert-guided motion-encoding search. The green boxes represent nodes, with each node representing the state of the car. The yellow box represents a driving intention, and each driving intention represents a structured sequence of actions. Figure(a)  represents the root node, where we illustrate three sampled driving intentions of the red vehicle for overtaking. The first intention involves making a probing maneuver to the left and eventually obtaining cooperation from the vehicle on the left, leading to Figure (b). Afterwards, the red vehicle can further consider continuing the lane change, eventually reaching node (e).The second intention involves directly changing lanes to the right without any probing action. Due to the lack of a probing maneuver, it does not allow enough reaction time for the vehicle on the right, resulting in a collision and leading to Figure(c). The comparison between these two intentions highlights the necessity of probing actions. The third intention involves accelerating first to create distance from the adjacent vehicles, resulting in Figure (d). Afterward, the red vehicle can safely change lanes to the right without requiring cooperation from other vehicles.}
    \label{fig:motivation}
\end{figure}

The long-term, tactical-level planning 
 in autonomous driving task leads to an exponential increase in the number of search nodes, resulting in intensive computational demands,  which limits the current applications of MCTS in the field of autonomous driving to near-field driving tasks\cite{9636442}, such as lane changing and following the preceding vehicle.

To alleviate this issue, \cite{hubert2021learning} proposed a sampling-based approach that computes an improved policy over a subset of the action space, reducing the number of traversed nodes during the search process. On the other hand, \cite{danihelka2021policy} employed the Gumbel Top-k trick, a method of non-repetitive sampling, to achieve more efficient search with fewer simulations.
Although these algorithms can enhance search efficiency to some extent, they are unable to address the exponential growth of node search problem in driving tasks caused by longer planning horizons.

Other research efforts have focused on the action space of driving tasks, discretizing the decision-making process for lane changing\cite{9147369} in scenarios involving straight-line driving. 
While this approach can address the complexity of search in driving tasks, it greatly restricts the driving modality and scenarios, excluded actions such as tentative lane changing in the driving policy.
This restriction renders it unsuitable for a wide range of driving tasks.

To address the long-horizon planning problem and the searching efficiency problem, in this paper we propose the expert-guided motion-encoding tree search (EMTS) algorithm.
By incorporating motion primitives, our approach enables each node to conduct searches based on a sequence of motions over a specific duration, rather than individual actions at each moment. This approach allows us to reduce the search complexity for long-duration MCTS tasks by decreasing the depth of the search tree.
EMTS compress the motion primitives into a comprehensive motion primitive latent space for sampling and expert policies encoding with an autoencoder architecture.
Thus enables the vehicle has the potential to sample any possible trajectories instead of restricted motions and enables long-horizon planning.
Moreover, expert policies are incorporate into EMTS as sampling candidates in the search phase and as the prior in the Bayesian process of the training phase to improve the searching efficiency of EMTS.
The pipeline of the proposed EMTS is shown at Fig. \ref{fig:motivation}.
As shown in Fig. \ref{fig:motivation} (b) and (e), the proposed EMTS can use tentative lane changing motion to interact with other vehicle.
And as shown in Fig. \ref{fig:motivation} (c), (d) and (f), the proposed EMTS can compare different actions to select the feasible action.

In summary, our contributions are as follows:
\begin{itemize}
\item We incorporate motion primitive methods into MCTS to enable long-term planning.
\item We leverage expert policies to maintain a multi-modal policy distribution and to improve searching efficiency.
\item We demonstrate the efficiency and effectiveness of our approach by comparing the proposed EMTS with other MuZero-based methods in three challenging scenarios.
\end{itemize}



\section{Related Works}

\subsection{MCTS in Reinforcement Learning}

The combination of MCTS and reinforcement learning initially made groundbreaking advancements in the realm of board games\cite{silver2016mastering, silver2017mastering}. AlphaGo\cite{silver2016mastering}, through pre-training on expert data and subsequent self-play, achieved a level of play that surpassed human capabilities in the game of Go. AlphaZero\cite{silver2017mastering}, on the other hand, broke free from dependence on pre-training and attained higher autonomy and independence. Simultaneously, it underwent simplification and optimization in its algorithmic structure, enabling its applicability to a wider range of board games.

The MuZero series\cite{Schrittwieser2019MasteringAG, ye2021mastering, ozair2021vector, antonoglou2021planning} extends the application of tree search methods to a wider range of scenarios, even in cases where the dynamic model of the environment are unknown. 
\cite{Schrittwieser2019MasteringAG} firstly maps observations to a latent space and learns an environment transition model within this latent space, then utilizes this model for prediction and planning. \cite{hubert2021learning} proposed a sampling-based approach, enabling MCTS to be applicable in continuous action or high-dimensional discrete action spaces. These algorithms do not rely on expert data but instead require a higher number of interactions, which can be quite costly for driving tasks. \cite{yin2022planning} utilizes extra rewards to assist the search process by providing action sequences similar to expert data. However, it still faces challenges in handling the selection of multimodal actions.

\subsection{Motion Primitives}
Motion primitive methods provide an effective approach for robot motion planning and control, by decomposing complex motions into basic motion patterns.
Optimization-based methods \cite{howard2007optimal, liu2017search, liu2018towards} allow for obtaining trajectories with various motion modes by adjusting the objective function, but is susceptible to getting trapped in local optima. 
Geometric based methods, such as Dubbins, Redsshep or 5-order polinomial curves, obtain the simplest and most efficient paths, but they are limited by the representation of predifiend shapes.
Parametric-based methods \cite{lucas2012bertrand} can provide smooth paths that adhere to the kinematics of the vehicle. However, they may struggle to accurately represent complex shapes.

In general, utilizing a variety of motion primitive methods can effectively represent a wider range of trajectory modes. However, there is no unified parameter representation or distribution pattern among different methods. Our previous work \cite{zhou2022accelerating} proposed the TaEc Motion Skill Distillation method to unify the representation of different motion primitive trajectories and utilized trajectory-level exploration instead of raw actions. However, the problem of longer horizon planning for combining multiple sequences of skills has not yet been resolved.

\section{METHODOLOGY}
We consider the standard formulation of RL, which is a Markov Decision Process defined by the tuple $\{\mathcal{S}, \mathcal{A}, \mathcal{T}, \mathcal{R}, \gamma\}$ of states, actions, transition probabilities, rewards, and discount factor.
The discounted cumulative return is defined as $G_t:=\sum_{i=t}^M\gamma_{i-t}r(s_t, a_t)$. The goal of RL is to maximize the expected return 
$\mathbb{E}_{s_0 \sim S_0}[G_0|s_0]$.

Our method consists of three parts:
\textbf{Comprehensive Motion Primitive Latent Space Construction}:
We leverage a skill distilling method to model low-level action sequences as comprehensive parameterized skill features.
\textbf{Expert Intention Encoding}:
We utilize expert data from different driving polices to extract expert action sequences, ensuring that even under the same observation, there are multiple potential polices to be chosen from.
\textbf{MCTS search with expert
policy involvement and skill-based strategy execution}: We have designed a new expert policies guided MCTS with a skill-level execution during the acting phase.

\begin{figure}[ht!]
    \centering
    \includegraphics[width=\columnwidth]{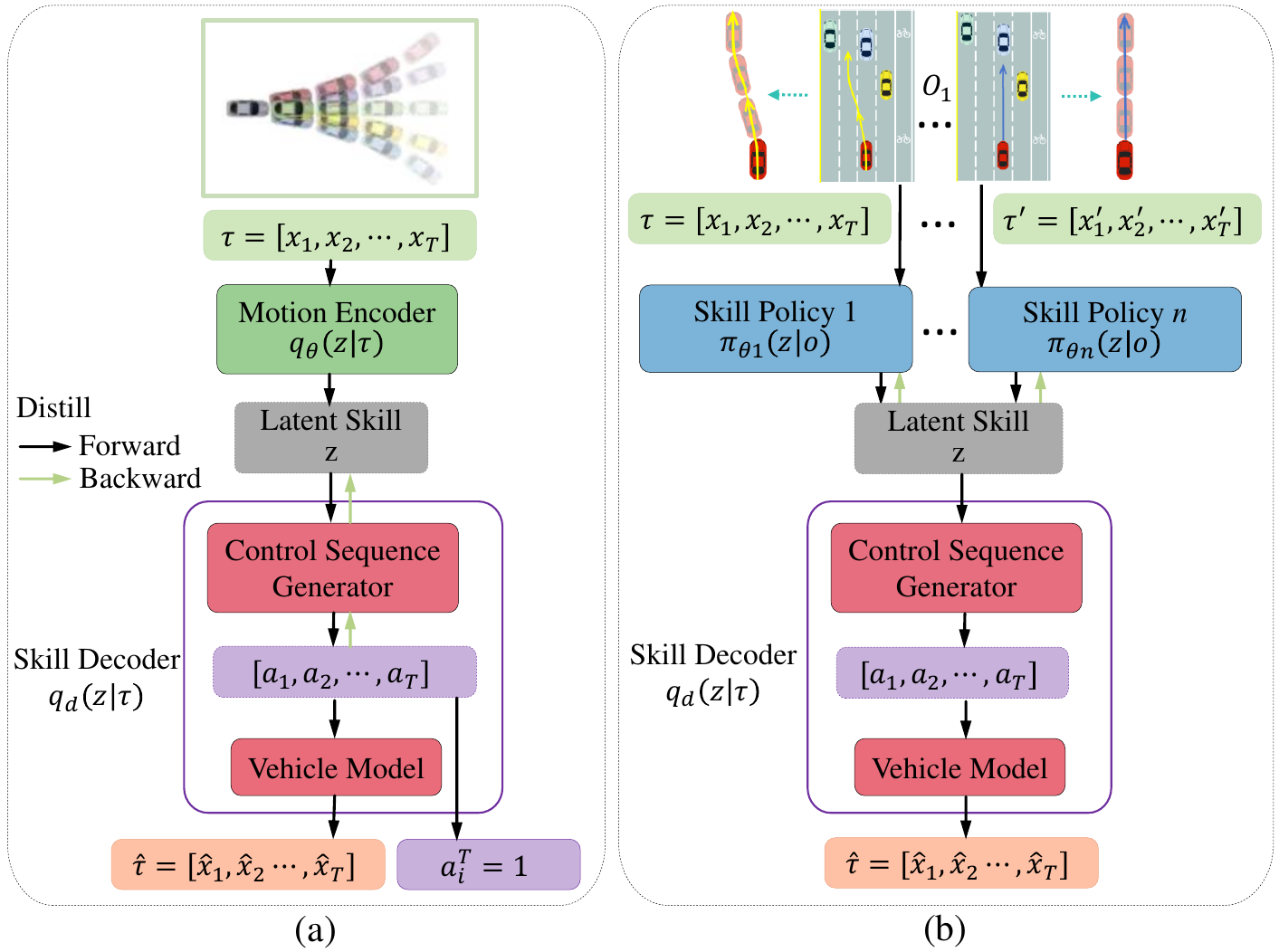}
    \caption{Illustration of Comprehensive Motion Primitive Latent Space Construction in (a), and Expert Intension Encoding in (b).
    In Figure (a), We first establish a Task Agnostic Ego Centric Library to encompass all the necessary skills required during driving. Then, we utilize the skill distillation method to provide a unified parameterized representation for various types of trajectories.In (b), we extract the policies from different expert data. Even when faced with the same observation, different intentions can lead to distinct sequential decision-making processes. We extract these policies into a latent space and utilize the decoder, which has already established a unified representation in step (a), to reconstruct trajectories. By employing this approach, the similarity of expert intentions can also be manifested through the distance in the latent space.}
    \label{fig:skill_expert_extraction}
\end{figure}

\subsection{Comprehensive Motion Primitive Latent Space} 




In this section, we refine and advance our previous work \cite{zhou2022accelerating} called Task Agnostic Ego Centric (TaEc) skill distillation.
Firstly, we establish a general-purpose library that encompasses a wide range of skills and can be utilized across various tasks. This is achieved by exhaustively sampling trajectories using diverse motion primitive methods in both spatial and temporal dimensions.
Secondly, skill distillation is employed to obtain a comprehensive parameterized representation for these skills,
as shown in Fig \ref{fig:skill_expert_extraction} (a).  It primarily trains a generative model that includes a motion encoder $q_m(z|\mathbf{\tau})$, which compressing trajectory information into low-dimensional latent space, and a skill decoder $q_d(\mathbf{\tau}|z)$, which reconstructing the trajectory from the latent space. 
The training of this model utilizes the evidence lower bound (ELBO) as follows:
\begin{equation}\label{eq:ELBO}
    \mathbb{E}_{q_m} \bigg[\underbrace{\log q_d(\tau \vert z)}_{\text{reconstruction}} - \zeta \big(\underbrace{\log q_m(z \vert \tau) - \log p(z)}_{\text{regularization}}\big) \bigg],
\end{equation}
The output $\tau$ of the decoder has two forms. 
One form is a sequence of actions, $\mathbf{[a_1, a_2, ..., a_T]}$, where $\mathbf{a_i} = [a_{throttle}, a_{steer}], for i \in {1,2,...T}$ denotes the throttle-steer action at timestep $i$.
In the given initial state and with the knowledge of the vehicle's kinematic model, we can obtain the other output from the decoder:
$\mathbf{[x_1, x_2, ..., x_T]}$, where $\mathbf{x_i} = [x_i, y_i, \theta_i, v_i], for  i \in {1,2,...T}$ denotes the future position, orientation and velocity of the vehicle.
Since the vehicle's kinematic model is deterministic and known, these two output modes are equivalent. 
During the phase of expert intention encoding, we use the former output mode.
During the phase of MCTS plan and execution, we use the latter output mode.

On one hand, the construction of the comprehensive motion primitive latent space encodes all the skills that a vehicle might use into the latent space, making our representation of skills sufficiently flexible and compact. On the other hand, the similarity of skills is reflected through the distance in the latent space, providing a certain regularity in the distribution of skills.


\subsection{Expert Intention Encoding}


Since the unavailability of expert policies directly, we choose to learn from different types of expert demonstrations. Given different expert datasets $ \{D_n^E\}_N = \{D_1^E, ..., D_N^E\}$,
where N is the number of datasets, for each dataset $ D_n^E \mathrel{:} \{ \xi_i \}$, and $\xi_i = (o_0, a_0, o_1, a_1, ..., o_{T_h}, a_{T_h}) $ denotes the observation-action pairs. In autonomous driving filed, the observation encompasses not only the ego vehicle's pose but also the information regarding surrounding vehicles and navigation. And action demotes the steer-throttle control.

The expert skill extraction pipeline is shown in \ref{fig:skill_expert_extraction} (b).
For an observation-action sequence with a length of T, we only use the information from the first frame $o_1$ and the future action sequences at T time steps $\tau= {[a_1, a_2, ..., a_T]}$ as inputs for intention encoding.
For each expert policy parameterized by $\theta_n$, we can train the skills by minimizing the mean squared error (MSE) of the trajectories to optimize the expert policy network.
\begin{equation}\label{eq:ELBO}
    \mathbb{E}_{o_1, \tau \sim D_n^E}\lVert q_d^{\perp}(\pi_{\theta_n(o_1)}) - \tau \rVert^2
\end{equation}
where $\cdot^{\perp}$ denotes that the operation does not participate in backpropagation.
It is worth noted that we freeze the parameter of the skill decoder network and train the encoder network, which mapping the states to the latent space z. 
In this way, the expert trajectories is encoded as latent skills in a unchanged motion primitives latent space.
Considered that the expert trajectories represent different driving intentions, the aforementioned process learns these intentions.

This approach allows us to understand and simulate the decision-making process of drivers from a high-level perspective, rather than simply replicating their specific low-level control actions. 
By learning driving intentions, we can adapt more flexibly to different driving scenarios and have a certain level of generalization ability.

\begin{figure*}[t]
    \centering
    \includegraphics[width=\textwidth]{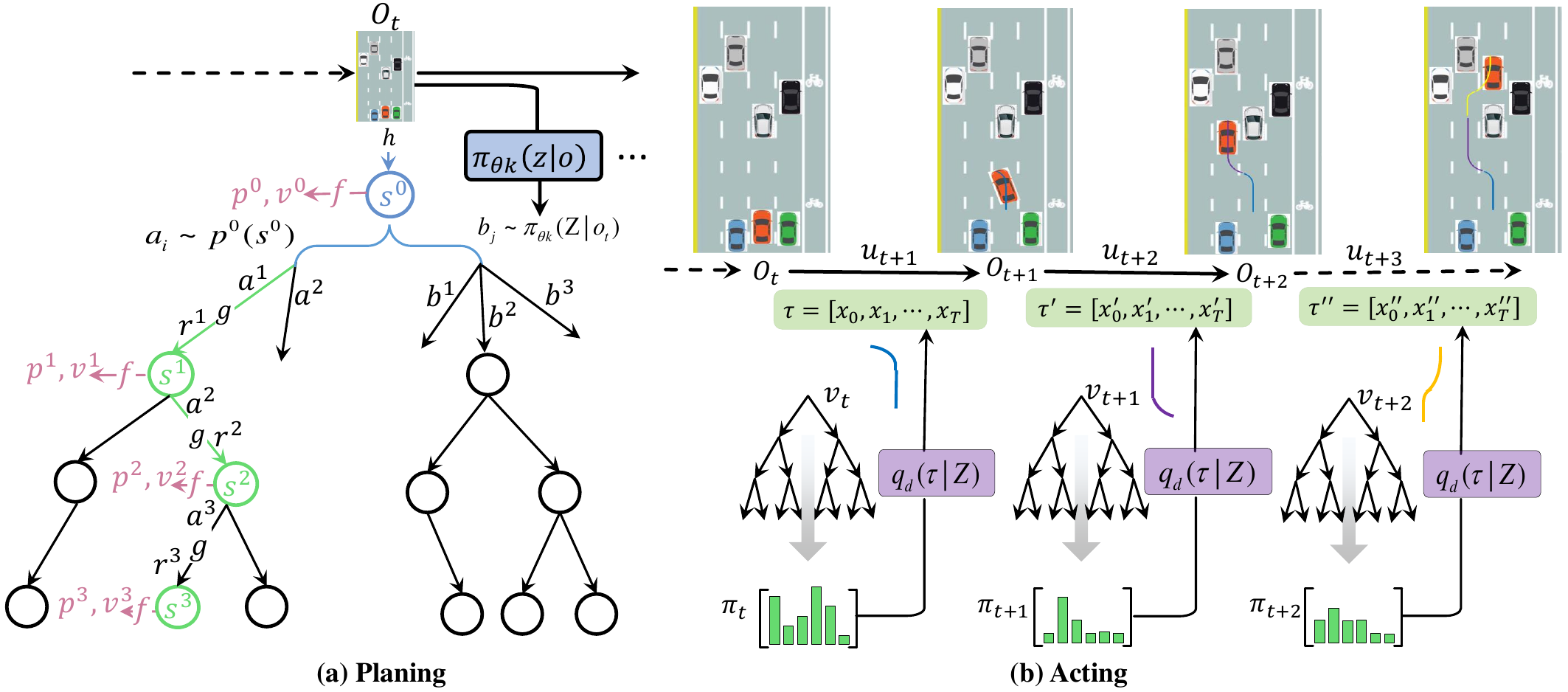}
    \caption{The pipeline of planning and action in MCTS. (a) The search process of MCTS. The sampling policy at the root search node of MCTS consists of two parts: the first part is the policy sampled by the prediction network of the MuZero network based on the root node $s_0$; the other part is composed of several expert policies sampled separately based on the original observation $o_t$.During the node expansion phase, getting closer to a certain expert policy will also gain extra rewards in the early stages of training to encourage exploration.Meanwhile, we adopt a Bayesian posterior based on expert data prior and visit count observations as the improved policy, rather than the visit count itself. (b), the execution process of MCTS, each time an intention z is output, we will obtain a series of trajectories $\tau$ through the $q_d$, and execute this series of trajectories within the future T time.}
    \label{fig:pipeline}
\end{figure*}


For different expert strategies, they share the same fixed skill decoder because the skill decoder has already established the normalized parameters for all possible trajectories. 
Therefore, the similarity between different expert strategies can also be reflected by the proximity in the latent space. 
By mapping the actions of different experts onto the latent space, we can observe the clustering or distribution patterns, which can provide insights into the similarities and differences among their driving behaviors. 
This allows us to better understand the underlying factors that contribute to different driving strategies and potentially identify common patterns or principles that can be applied across various driving scenarios.

\subsection{MCTS search with expert
policy involvement and skill-based strategy execution}
\subsubsection{Model}

Our Monte Carlo network architecture is inherited from the Sampled MuZero \cite{hubert2021learning} and consists of the following three components:
\begin{equation}\label{eq:Muzero}
    \begin{aligned}
    s_0 = h_{\theta}(o_1, ..., o_t) \\
    r_k, s_k = g_{\theta}(s_{k-1}, z_k) \\
    \mathbf{p_k}, v_k = f_{\theta}(s_k)
    \end{aligned}
\end{equation}
$h_{\theta}$ is the representation network, which maps the observation $o_t$ into hidden state $s_t$, and perform subsequent predictions in the hidden space during the search. 
Since we conduct the search in the
skill space instead of the raw action space, the dynamic network $g_{\theta}$ takes the current node's hidden state $s_{k-1}$ and skill variable $z_k$ as inputs to predict the resulting hidden state $s_k$ and cumulative reward $r_k$ after executing the entire sequence of actions associated with the skill.
The prediction network, $f_{\theta}$ takes the hidden state as input and predicts the value $v_k$ and skill policy $\mathbf{p_k}$. In our network, the policy $\mathbf{p_k} = \sum_{i=1}^M \alpha_i \mathcal{N}(\mu, \sigma^2)$ is the Gaussian Mixture Model.
This mixture density network allows the skill to be a distribution with multiple peaks, representing multiple selectable strategies that can be sampled or expanded.

In addition to Monte Carlo network, the expert encoder and skill decoder are used separately in search phase and acting phase.
The overall process can be illustrated as shown in Fig\ref{fig:pipeline}. 

\subsubsection{Search}

In the \textbf{sampling phase} of the root node, in addition to the samples generated by the original policy of MuZero, we also sample skills from different expert policies as additional components of the child nodes, as shown in Fig \ref{fig:pipeline} (a).
To be more precise, we employ a sampling approach that relies on the following mixture policy.
\begin{equation}\label{eq:Muzero}
\overline{\pi}_{sample} = \alpha * \sum_{i=0}^{N-1}\pi_{\theta i}(z|o) / N + (1-\alpha) * f_{\theta}(z|s)
\end{equation}
This expansion procedure with mixture policy  only exists when sampling children of the root nodes. 
This approach ensures that the expert policies independent from the representation network, allowing the expert data to be trained only once without the need for a separate expert buffer as in \cite{yin2022planning}. 
In addition, this approach avoids the cumulative model error problem associated with increasing tree depth during MuZero training \cite{liu2023efficient}.

During the \textbf{expansion phase} of the nodes, we also incorporate additional probability boosts for skills that are close to expert policies. 
The UCB formula augmented with expert policy is as follows:
\begin{equation}\label{eq:Muzero}
    max_z Q(s, z) + P_{adj}(s, z)[c1 + log(\frac{\sum_b N(s,b) + c2 + 1}{c2}]
\end{equation}
where $P_{adj}(s, z)$ can be calculated by:

\begin{equation}\label{eq:Muzero}
\adjustbox{max width=\columnwidth}{
$    
P_{adj}(s, z) = 
    \begin{cases}
    \gamma * max_i \pi_{\theta i}(z|o_t) + (1 - \gamma) * f(z|s) &\text{if } s=s_0, \\
    f(z|s) & \text{else } 
    \end{cases} 
$

}    
\end{equation}


This enables skills that are closer to any kind of expert policies to be more easily selected and expanded, thereby enhancing the effectiveness of searching near expert policies. 
As training progresses, $\gamma$ gradually decreases, indicating that the additional probability boosts diminish.
By gradually reducing the reliance on expert data, we can uncover novel and superior strategies that may not have been captured by the expert policies alone. 

\subsubsection{Acting}
During the Acting process, we select a skill based on the normalized probability of the visit count from the root child nodes. 
As shown is Fig\ref{fig:pipeline} (b), once a specific skill sample is chosen, it is passed through the skill decoder network to obtain the action sequence $\mathbf{[a_1, a_2, ..., a_T]}$ for the subsequent T time steps. 
The vehicle executes the whole sequence of actions at once, during which it does not make any further decisions, and the sum of the rewards obtained is considered as the reward for executing the current skill.

\subsubsection{Training}
Our loss can be mainly divided into three parts:
\begin{equation}\label{eq:Muzero}
l_t(\theta) = \sum_{j=0}^J l^r(u_{t+j}, r_t^j) + l^v(z_{t+j}, v_t^j) + l^p(\pi_{t+j}, p_t^j) + c\lVert \theta \rVert^2
\end{equation}
where $u_{t+j}$ is the sum of rewards obtained after executing T consecutive  raw actions.
$z_{t+k}$ is the bootstrapped n-step target, and $\pi_{t+j}$ is the improved policy.
In order to maintain the diversity of policy outputs, we designed a bayesian inference method to calculate the posterior probability as the improved policy.
Let $\{z_k\}_K$ be K intentions from the sampled distribution.
The posterior probability $\pi$ can be calculated as:
\begin{equation}\label{eq:Muzero}
\pi(z_k|s) = \frac{(p_{ME}(z_k|o))^{1/\lambda} \cdot w(z_k|s)}{\sum_{k=1}^K (p_{ME}(z_k|o))^{1/\lambda} \cdot w(z_k|s)}
\end{equation}
where $w(z|s)$ and $p_{ME}(z|o)$ respectively represent the normalized visited count and fused expert prior distribution.
\begin{equation}\label{eq:Muzero}
w(z|s) = \frac{\exp(N^{(k)}(s, z))}{\sum_{k=1}^{K}\exp(N^{(k)}(s, z))}
\end{equation}
\begin{equation}\label{eq:Muzero}
p_{ME}(z|o) = \sum_{k=1}^K\sum_{n=1}^Nm_n\cdot \pi_{\theta n}(z|o)\cdot \delta(z-z_k)
\end{equation}
In this context, $N(s,z)$ represents the visit count of the sample $z_k$, and $m_n$ represents the weight of the expert policy, and $\lambda$ represents the temperature coefficient.

Different from MuZero, who uses the normalized visit count $w(z|s)$ as improved policy, we used the $\pi(z_k|s)$ as the improved policy. 
This mechanism maintains a probability for under-explored expert policy in its latent space, ensuring that MCTS preserves diverse enough intentions for future exploration, thereby preventing confinement to local optima.

\begin{figure}[t]
    \centering
    \includegraphics[width=0.8\columnwidth]{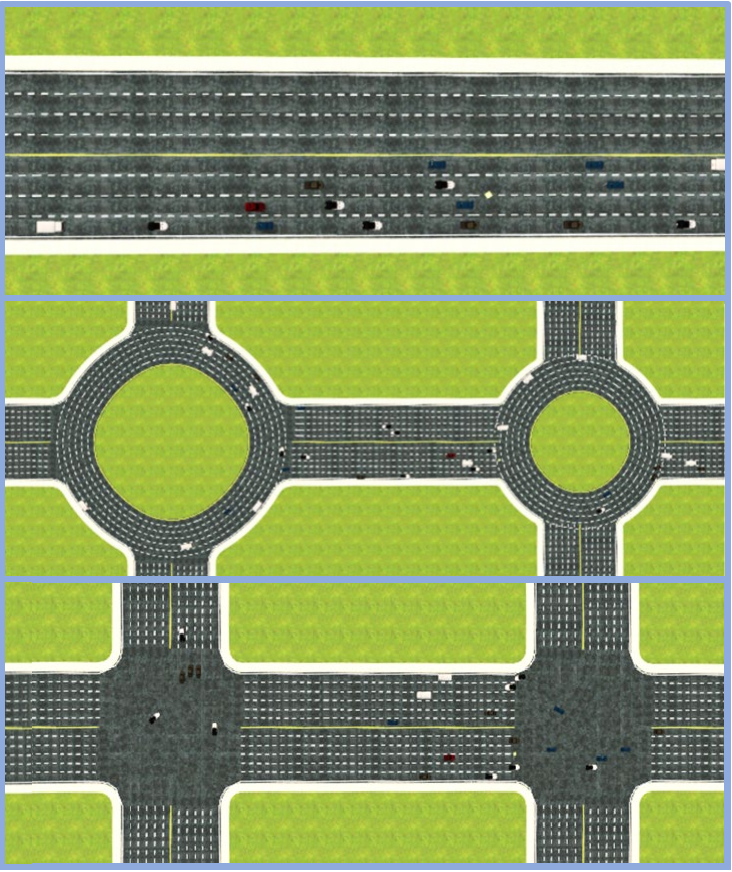}
    \caption{Three scenarios of our experiments listed from top to down: highway, roundabout and intersection. }
    \label{fig:scenarios}
\end{figure}

\begin{figure*}[t]
    \centering
    \includegraphics[width=\textwidth]{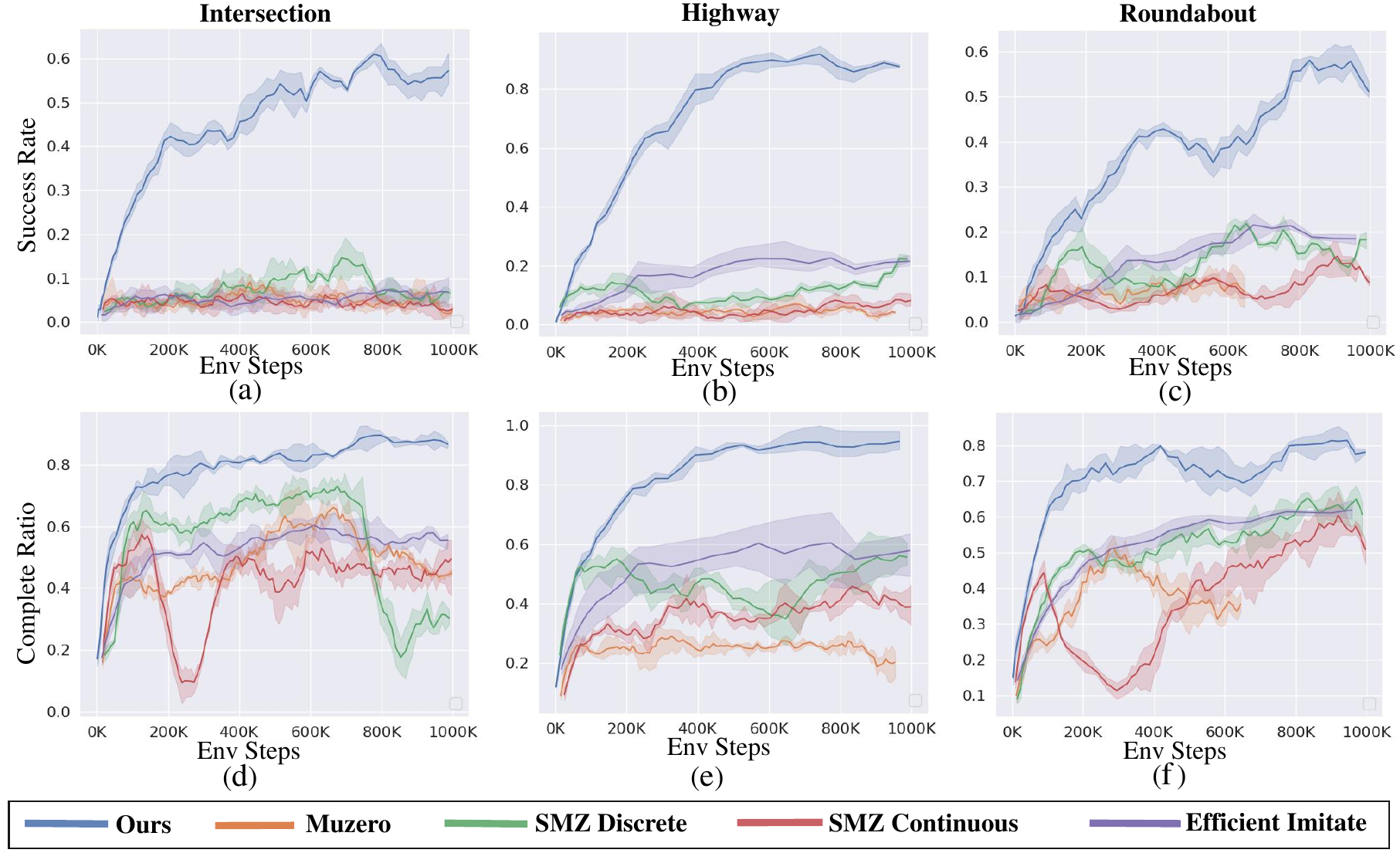}
    \caption{Comparison of our method with other baselines. The data shows that the MCTS algorithm with expert data, including Efficient Imitate and ours, has advantages in terms of training speed and effectiveness. Our algorithm outperforms other algorithms in various scenarios.}
    \label{fig:experiment_results}
\end{figure*}

\section{EXPERIMENT}

\subsection{Experiment Setting}
\subsubsection{Environment and task}
We utilized three scenarios from the MetaDrive simulator\cite{li2021metadrive}, namely intersection, highway, and roundabout, to validate our experiments. 
In the intersection scenario, we set the density rate to 0.45 to represent the highest traffic and perform tasks such as going straight, making right/left turn, and U-turn in an unprotected turn scenario.
In highway and roundabout scenario, we set the density rate to 0.3 and make a strict passing limit, in order to encourage overtaking behavior, as purely following other vehicles would lead to timeouts.
 We employed a 5-channel bird's-eye view image as the observation, following the approach outlined in \cite{zhou2022accelerating}.

\subsubsection{Reward and Step Information} 

The reward function consists of two components. 
The first component is the dense reward, which rewards the vehicle for moving forward and penalizes unstable driving behavior. 
The second component is the sparse reward, which provides a reward when the vehicle succeeds and penalizes timeouts and collisions.

The result reward function is as follows:
\begin{equation} \label{eq:reward}
    r_t = c_1 \cdot R_{driving} + c_2 \cdot R_{speed} + c_3 \cdot R_{jerk} + c4 \cdot R_{termination} 
\end{equation}
The first three items represent dense rewards, while the last item represents a sparse reward.
\subsubsection{Baselines} We compare the performance of the proposed method with several baselines:
\begin{itemize}
    \item \textbf{Muzero}\cite{Schrittwieser2019MasteringAG}: 
    Due to the discrete action space requirement of MuZero, we discretized the throttle and steering inputs into seven dimensions each. As a result, the algorithm performs its search in a 49-dimensional action space.
    \item \textbf{Sampled Muzero Discrete}: Following the approach described in \cite{hubert2021learning}, we sample K actions from the aforementioned discrete action space to increase the efficiency of the search process. 
    \item \textbf{Sampled Muzero Continous}\cite{hubert2021learning}: We sample K actions from the continuous action space, following a policy that conforms to a Gaussian distribution.
    \item \textbf{Efficient Imitate} \cite{yin2022planning}: An approach to assist Monte Carlo Tree Search with expert data.
\end{itemize}
For the first three baseline algorithms, we replicated them using LightZero \cite{lightzero}. As for the last one, we referred to \cite{yin2022planning} for guidance.
\subsubsection{Evaluation} 
We employ success rate and completion rate as evaluation metrics, with each algorithm interacting with the environment for 1 million steps.

We set the sampling quantity as K=20, the number of nodes per search as 100, and the skill horizon as T=10. 
We've gathered data from three distinct experts, signified by N=3. Concurrently, we've configured the Gaussian Mixture Model to have three peaks, denoted by M=3.

\subsection{Experiment results}
The comparison results are shown in Fig. \ref{fig:experiment_results}.

In all of the three scenarios the proposed EMTS method achieves the best performance.
EMTS achieves the peak success rate of approximately $60\%$, $90\%$, and $60\%$ in the three experiment scenarios, respectively.
And the best peak success rate of other algorithms are approximately $20\%$, $30\%$, and $25\%$ in these three experiment scenarios, respectively.
In these three experiment scenarios, the SMZ discrete has the best performance in the Intersection and the Roundabout scenarios and efficient imitate algorithm has the best performance in the Highway scenario. 
If can also be observed that the EMTS method is the most efficient algorithm at the early training stage.
As shown in Fig. \ref{fig:skill_expert_extraction}, the Intersection scenario exhibits a higher traffic density and shared space positions for traffic from different directions. 

In terms of completion ratio, our algorithm reaches nearly $100\% $completion rate around 500k steps in the Intersection scenario. 
Most algorithms achieve a success rate of over $60\%$. SMZ Continuous exhibits a rapid decline and subsequent increase after 200k steps. 
In the Highway scenario, our algorithm maintains a completion rate of $100\% $throughout. 
Efficient Imitate quickly rises to over $60\%$ completion rate around 200k steps, with some minor fluctuations thereafter. 
The completion ratios of the other three algorithms remain below $50\%$.
In the Roundabout scenario, our algorithm reaches a peak completion rate of $80\%$ around 400k steps, with some oscillation afterwards. Efficient Imitate and SMZ Discrete show similar performance, achieving around $60\%$ completion rate at around 800k steps. 
SMZ Continuous experiences a rapid decline after an initial period of rapid ascent and continues to slowly rise until around 250k steps.

Our algorithm exhibits significant advantages in both metrics across the three scenarios.

\section{CONCLUSIONS}

In this paper, we proposed the expert-guided motion-encoding tree search (EMTS) algorithm to deal with the search efficiency problem in long-horizon planning.
EMTS adopt the comprehensive motion primitive latent space to reduce the search depth while maintaining the skill diversity.
In addition EMTS also leverage expert policy to maintain multi-modal policy distribution to improve the search efficiency.
The experiment results demonstrate that the proposed EMTS algorithm surpasses other four baseline algorithms in both success rate and complete ratio.
For future work, we may focus on implementing the proposed EMTS in other challenging applications such as manipulator motion planning and dense pedestrian navigation.

\addtolength{\textheight}{-12cm}   




\bibliographystyle{IEEEtran}
\bibliography{ref.bib}

\end{document}